# Acquiring Word-Meaning Mappings
# for Natural Language Interfaces

**Cynthia A. Thompson**                                    CINDI@CS.UTAH.EDU
*School of Computing, University of Utah*
*Salt Lake City, UT 84112-3320*

**Raymond J. Mooney**                                    MOONEY@CS.UTEXAS.EDU
*Department of Computer Sciences, University of Texas*
*Austin, TX 78712-1188*

## Abstract

This paper focuses on a system, WOLFIE (WOrd Learning From Interpreted Examples), that acquires a semantic lexicon from a corpus of sentences paired with semantic representations. The lexicon learned consists of phrases paired with meaning representations. WOLFIE is part of an integrated system that learns to transform sentences into representations such as logical database queries.

Experimental results are presented demonstrating WOLFIE's ability to learn useful lexicons for a database interface in four different natural languages. The usefulness of the lexicons learned by WOLFIE are compared to those acquired by a similar system, with results favorable to WOLFIE. A second set of experiments demonstrates WOLFIE's ability to scale to larger and more difficult, albeit artificially generated, corpora.

In natural language acquisition, it is difficult to gather the annotated data needed for supervised learning; however, unannotated data is fairly plentiful. Active learning methods attempt to select for annotation and training only the most informative examples, and therefore are potentially very useful in natural language applications. However, most results to date for active learning have only considered standard classification tasks. To reduce annotation effort while maintaining accuracy, we apply active learning to semantic lexicons. We show that active learning can significantly reduce the number of annotated examples required to achieve a given level of performance.

## 1. Introduction and Overview

A long-standing goal for the field of artificial intelligence is to enable computer understanding of human languages. Much progress has been made in reaching this goal, but much also remains to be done. Before artificial intelligence systems can meet this goal, they first need the ability to *parse* sentences, or transform them into a representation that is more easily manipulated by computers. Several knowledge sources are required for parsing, such as a grammar, lexicon, and parsing mechanism.

Natural language processing (NLP) researchers have traditionally attempted to build these knowledge sources by hand, often resulting in brittle, inefficient systems that take a significant effort to build. Our goal here is to overcome this "knowledge acquisition bottleneck" by applying methods from machine learning. We develop and apply methods from *empirical* or *corpus-based* NLP to learn semantic lexicons, and from *active learning* to reduce the annotation effort required to learn them.

The semantic lexicon is one NLP component that is typically challenging and time consuming to construct and update by hand. Our notion of semantic lexicon, formally defined in Section 3, is that of a list of phrase-meaning pairs, where the meaning representation is determined by the language understanding task at hand, and where we are taking a compositional view of sentence meaning (Partee, Meulen, & Wall, 1990). This paper describes a system, Wolfie (WOrd Learning From Interpreted Examples), that acquires a semantic lexicon of phrase-meaning pairs from a corpus of sentences paired with semantic representations. The goal is to automate lexicon construction for an integrated NLP system that acquires both semantic lexicons and parsers for natural language interfaces from a single training set of annotated sentences.

Although many others (Sébillot, Bouillon, & Fabre, 2000; Riloff & Jones, 1999; Siskind, 1996; Hastings, 1996; Grefenstette, 1994; Brent, 1991) have presented systems for learning information about lexical semantics, we present here a system for learning lexicons of phrase-meaning pairs. Further, our work is unique in its combination of several features, though prior work has included some of these aspects. First, its output can be used by a system, Chill (Zelle & Mooney, 1996; Zelle, 1995), that learns to parse sentences into semantic representations. Second, it uses a fairly straightforward batch, greedy, heuristic learning algorithm that requires only a small number of examples to generalize well. Third, it is easily extendible to new representation formalisms. Fourth, it requires no prior knowledge although it can exploit an initial lexicon if provided. Finally, it simplifies the learning problem by making several assumptions about the training data, as described further in Section 3.2.

We test Wolfie's ability to acquire a semantic lexicon for a natural language interface to a geographical database using a corpus of queries collected from human subjects and annotated with their logical form. In this test, Wolfie is integrated with Chill, which learns parsers but requires a semantic lexicon (previously built manually). The results demonstrate that the final acquired parser performs nearly as accurately at answering novel questions when using a learned lexicon as when using a hand-built lexicon. Wolfie is also compared to an alternative lexicon acquisition system developed by Siskind (1996), demonstrating superior performance on this task. Finally, the corpus is translated into Spanish, Japanese, and Turkish, and experiments are conducted demonstrating an ability to learn successful lexicons and parsers for a variety of languages.

A second set of experiments demonstrates Wolfie's ability to scale to larger and more difficult, albeit artificially generated, corpora. Overall, the results demonstrate a robust ability to acquire accurate lexicons directly usable for semantic parsing. With such an integrated system, the task of building a semantic parser for a new domain is simplified. A single representative corpus of sentence-representation pairs allows the acquisition of both a semantic lexicon and parser that generalizes well to novel sentences.

While building an annotated corpus is arguably less work than building an entire NLP system, it is still not a simple task. Redundancies and errors may occur in the training data. A goal should be to also minimize the annotation effort, yet still achieve a reasonable level of generalization performance. In the case of natural language, there is frequently a large amount of unannotated text available. We would like to automatically, but intelligently, choose which of the available sentences to annotate.

We do this here using a technique called *active learning*. Active learning is a research area in machine learning that features systems that automatically select the most informative examples for annotation and training (Cohn, Atlas, & Ladner, 1994). The primary goal of active learning is to reduce the number of examples that the system is trained on, thereby reducing the example annotation cost, while maintaining the accuracy of the acquired information. To demonstrate the usefulness of our active learning techniques, we compared the accuracy of parsers and lexicons learned using examples chosen by active learning for lexicon acquisition, to those learned using randomly chosen examples, finding that active learning saved significant annotation cost over training on randomly chosen examples. This savings is demonstrated in the geography query domain.

In summary, this paper provides a new statement of the lexicon acquisition problem and demonstrates a machine learning technique for solving this problem. Next, by combining this with previous research, we show that an entire natural language interface can be acquired from one training corpus. Further, we demonstrate the application of active learning techniques to minimize the number of sentences to annotate as training input for the integrated learning system.

The remainder of the paper is organized as follows. Section 2 gives more background information on Chill and introduces Siskind's lexicon acquisition system, which we will compare to Wolfie in Section 5. Sections 3 and 4 formally define the learning problem and describe the Wolfie algorithm in detail. In Section 5 we present and discuss experiments evaluating Wolfie's performance in learning lexicons in a database query domain and for an artificial corpus. Next, Section 6 describes and evaluates our use of active learning techniques for Wolfie. Sections 7 and 8 discuss related research and future directions, respectively. Finally, Section 9 summarizes our research and results.

## 2. Background

In this section we give an overview of Chill, the system that our research adds to. We also describe Jeff Siskind's lexicon acquisition system.

### 2.1 Chill

The output produced by Wolfie can be used to assist a larger language acquisition system; in particular, it is currently used as part of the input to a parser acquisition system called Chill (Constructive Heuristics Induction for Language Learning). Chill uses *inductive logic programming* (Muggleton, 1992; Lavrač & Džeroski, 1994) to learn a deterministic shift-reduce parser (Tomita, 1986) written in Prolog. The input to Chill is a corpus of sentences paired with semantic representations, the same input required by Wolfie. The parser learned is capable of mapping the sentences into their correct representations, as well as generalizing well to novel sentences. In this paper, we limit our discussion to Chill's ability to acquire parsers that map natural language questions directly into Prolog queries that can be executed to produce an answer (Zelle & Mooney, 1996). Following are two sample queries for a database on U.S. geography, paired with their corresponding Prolog query:

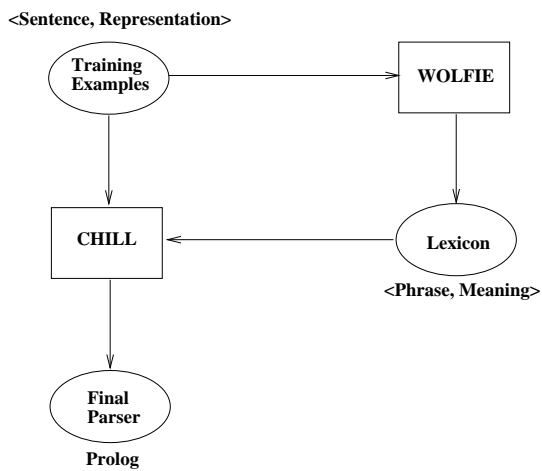

Figure 1: The Integrated System

What is the capital of the state with the biggest population?
```
answer(C, (capital(S,C), largest(P, (state(S), population(S,P))))).
```

What state is Texarkana located in?
```
answer(S, (state(S), eq(C,cityid(texarkana,_)), loc(C,S))).
```

CHILL treats parser induction as the problem of learning rules to control the actions of a shift-reduce parser. During parsing, the current context is maintained in a stack and a buffer containing the remaining input. When parsing is complete, the stack contains the representation of the input sentence. There are three types of operators that the parser uses to construct logical queries. One is the introduction onto the stack of a predicate needed in the sentence representation due to a phrase's appearance at the front of the input buffer. These operators require a semantic lexicon as background knowledge. For details on this and the other two parsing operators, see Zelle and Mooney (1996). By using WOLFIE, the lexicon is provided automatically. Figure 1 illustrates the complete system.

## 2.2 Jeff Siskind's Lexicon Learning Research

The most closely related previous research into automated lexicon acquisition is that of Siskind (1996), itself inspired by work by Rayner, Hugosson, and Hagert (1988). As we will be comparing our system to his in Section 5, we describe the main features of his research in this section. His goal is one of cognitive modeling of children's acquisition of the lexicon, where that lexicon can be used for both comprehension and generation. Our goal is a machine learning and engineering one, and focuses on a lexicon for comprehension and use in parsing, using a learning process that does not claim any cognitive plausibility, and with the goal of learning a lexicon that generalizes well from a small number of training examples.

His system takes an incremental approach to acquiring a lexicon. Learning proceeds in two stages. The first stage learns *which* symbols in the representation are to be used in the

```
(''capital'', capital(_,_)),          (''state'', state(_)),
(''biggest'', largest(_,_)),          (''in'', loc(_,_)),
(''highest point'', high_point(_,_)), (''long'', len(_,_)),
(''through'', traverse(_,_)),         (''capital'', capital(_)),
(''has'', loc(_,_))
```

Figure 2: Sample Semantic Lexicon

final "conceptual expression" that represents the meaning of a word, by using a version-space approach. The second stage learns *how* these symbols are put together to form the final representation. For example, when learning the meaning of the word "raise", the algorithm may learn the set {CAUSE, GO, UP} during the first stage and put them together to form the expression CAUSE($x$, GO($y$, UP)) during the second stage.

Siskind (1996) shows the effectiveness of his approach on a series of artificial corpora. The system handles noise, lexical ambiguity, referential uncertainty, and very large corpora, but the usefulness of lexicons learned is only compared to the "correct," artificial lexicon. The goal of the experiments presented there was to evaluate the correctness and completeness of learned lexicons. Earlier work (Siskind, 1992) also evaluated versions of his technique on a quite small corpus of real English and Japanese sentences. We extend that evaluation to a demonstration of the system's usefulness in performing real world natural language processing tasks, using a larger corpus of real sentences.

## 3. The Lexicon Acquisition Problem

Although in the end our goal is to acquire an entire natural language interface, we currently divide the task into two parts, the lexicon acquisition component and the parser acquisition component. In this section, we discuss the problem of acquiring semantic lexicons that assist parsing and the acquisition of parsers. The training input consists of natural language sentences paired with their meaning representations. From these pairs we extract a lexicon consisting of phrases paired with their meaning representations. Some training pairs were given in the previous section, and a sample lexicon is shown in Figure 2.

### 3.1 Formal Definition

To present the learning problem more formally, some definitions are needed. While in the following we use the terms "string" and "substring," these extend straight-forwardly to natural language sentences and phrases, respectively. We also refer to labeled trees, making the assumption that the semantic meanings of interest can be represented as such. Most common representations can be recast as labeled trees or forests, and our formalism extends easily to the latter.

**Definition:** Let $\Sigma_V$, $\Sigma_E$ be finite alphabets of vertex labels and edge labels, respectively. Let $V$ be a finite nonempty set of vertices, $l$ a total function $l : V \rightarrow \Sigma_V$, $E$ a set of unordered pairs of distinct vertices called edges, and $a$ a total function $a : E \rightarrow \Sigma_E$. $G = (V, l, E, a)$ is a *labeled graph*.

**String $s_1$: "The girl ate the pasta with the cheese.''**

**Tree $t_1$**

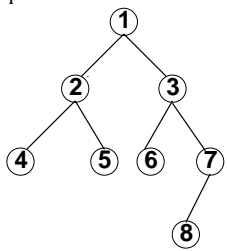

**$t_1$ with its vertex and edge labels:**

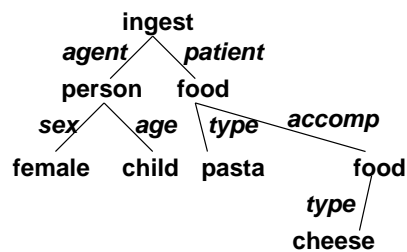

**Interpretation $f_1$ from $s_1$ to $t_1$ :**

> $f_1$("girl") = 2
> $f_1$("ate") = 1
> $f_1$("pasta") = 3
> $f_1$("the cheese") = 7

Figure 3: Labeled Trees and Interpretations

**Definition:** A *labeled tree* is a connected, acyclic labeled graph.

Figure 3 shows the labeled tree $t_1$ (with vertices 1-8) on the left, with associated vertex and edge labels on the right. The function $l$ is:[1]

> { (1, `ingest`), (2, `person`), (3, `food`), (4, `female`), (5, `child`), (6, `pasta`),
> (7, `food`), (8, `cheese`) }.

The tree $t_1$ is a semantic representation of the sentence $s_1$: "The girl ate the pasta with the cheese." Using a conceptual dependency (Schank, 1975) representation in Prolog list form, the meaning is:

```
[ingest,   agent:[person, sex:female, age:child],
             patient:[food, type:pasta, accomp:[food, type:cheese]]].
```

**Definition:** A *u-v path* in a graph $G$ is a finite alternating sequence of vertices and edges of $G$, in which no vertex is repeated, that begins with vertex $u$ and ends with vertex $v$, and in which each edge in the sequence connects the vertex that precedes it in the sequence to the vertex that follows it in the sequence.

**Definition:** A *directed, labeled tree* $T = (V, l, E, a)$ is a labeled tree whose edges consist of ordered pairs of vertices, with a distinguished vertex $r$, called the root, with the property that for every $v \in V$, there is a directed $r$-$v$ path in $T$, and such that the underlying undirected unlabeled graph induced by $(V, E)$ is a connected, acyclic graph.

**Definition:** An *interpretation* $f$ from a finite string $s$ to a directed, labeled tree $t$ is a one-to-one function mapping a subset $s'$ of the substrings of $s$, such that no two strings in $s'$ overlap, into the vertices of $t$ such that the root of $t$ is in the range of $f$.

---

1. We omit enumeration of the function $e$ but it could be given in a similar manner, for example ((1,2), `agent`) is an element of $e$.

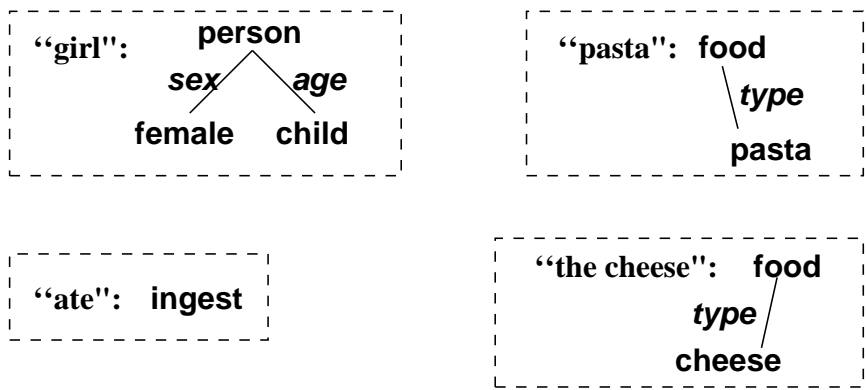

Figure 4: Meanings

The interpretation provides information about what parts of the meaning of a sentence originate from which of its phrases. In Figure 3, we show an interpretation, $f_1$, of $s_1$ to $t_1$. Note that "with" is not in the domain of $f_1$, since $s'$ is a subset of the substrings of $s$, thus allowing some words in $s$ to have no meaning. Because we disallow overlapping substrings in the domain, both "cheese" and "the cheese" could not map to vertices in $t_1$.

**Definition:** Given an interpretation $f$ of string $s$ to tree $t$, and an element $p$ of the domain of $f$, the *meaning* of $p$ relative to $s$, $t$, $f$ is the connected subgraph of $t$ whose vertices include $f(p)$ and all its descendents *except* any other vertices in the range of $f$ and their descendents.

Meanings in this sense concern the "lowest level" of phrasal meanings, occurring at the terminal nodes of a semantic grammar, namely the entries in the semantic lexicon. The grammar can then be used to construct the meanings of longer phrases and entire sentences. This is our motivation for the previously stated constraint that the root must be included in the range of $f$: we want all vertices in the sentence representation to be included in the meaning of some phrase. Note that the meaning of $p$ is also a directed tree with $f(p)$ as its root. Figure 4 shows the meanings of each phrase in the domain of interpretation function $f_1$ shown in Figure 3. We show only the labels on the vertices and edges for readability.

**Definition:** Given a finite set $STF$ of triples $< s_1, t_1, f_1 >, \ldots, < s_n, t_n, f_n >$, where each $s_i$ is a finite string, each $t_i$ is a directed, labeled tree, and each $f_i$ is an interpretation function from $s_i$ to $t_i$, let the *language* $\mathcal{L}_{STF} = \{p_1, \ldots, p_k\}$ of $STF$ be the union of all substrings[2] that occur in the domain of some $f_i$. For each $p_j \in \mathcal{L}_{STF}$, the *meaning set* of $p_j$, denoted $M_{STF}(p_j)$,[3] is the set of all meanings of $p_j$ relative to $s_i, t_i, f_i$ for some $< s_i, t_i, f_i > \in STF$. We consider two meanings to be the same if they are isomorphic trees taking labels into account.

For example, given sentence $s_2$: "The man ate the cheese," the labeled tree $t_2$ pictured in Figure 5, and $f_2$ defined as: $f_2$("ate") = 1, $f_2$("man") = 2, $f_2$("the cheese") = 3; the

---

2. We consider two substrings to be the same string if they contain the same characters in the same order, irrespective of their positions within the larger string in which they occur.

3. We omit the subscript on $M$ when the set $STF$ is obvious from context.

**String $s_2$ :  ''The man ate the cheese.''**

**Tree $t_2$ :**                                    **$t_2$ with its vertex and edge labels:**

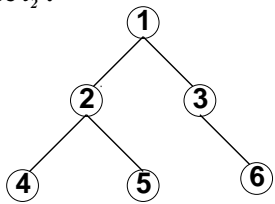         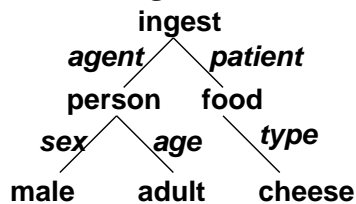

Figure 5: A Second Tree

meaning set of "the cheese" with respect to $STF = \{<s_1, t_1, f_1>, <s_2, t_2, f_2>\}$ is $\{[\texttt{food}, \texttt{type:cheese}]\}$, just one meaning though $f_1$ and $f_2$ map "the cheese" to different vertices in the two trees, because the subgraphs denoting the meaning of "the cheese" for the two functions are isomorphic.

**Definition:** Given a finite set $STF$ of triples $<s_1, t_1, f_1>, \ldots, <s_n, t_n, f_n>$, where each $s_i$ is a finite string, each $t_i$ is a directed, labeled tree, and each $f_i$ is an interpretation function from $s_i$ to $t_i$, the *covering lexicon* expressed by $STF$ is

$$\{(p, m) : p \in \mathcal{L}_{STF}, m \in M(p)\}.$$

The covering lexicon $L$ expressed by $STF = \{<s_1, t_1, f_1>, <s_2, t_2, f_2>\}$ is:

> { ("girl", [`person, sex:female, age:child`]),
> ("man", [`person, sex:male, age:adult`]),
> ("ate", [`ingest`]),
> ("pasta", [`food, type:pasta`]),
> ("the cheese", [`food, type:cheese`]) }.

The idea of a covering lexicon is that it provides, for each string (sentence) $s_i$, a meaning for some of the phrases in that sentence. Further, these meanings are trees whose labeled vertices together include each of the labeled vertices in the tree $t_i$ representing the meaning of $s_i$, with no vertices duplicated, and containing no vertices not in $t_i$. Edge labels may or may not be included, since the idea is that some of them are due to syntax, which the parser will provide; those edges capturing lexical semantics are in the lexicon. Note that because we only include in the covering lexicon phrases (substrings) that are in the domains of the $f_i$'s, words with the empty tree as meaning are not included in the covering lexicon. Note also that we will in general use "phrase" to mean substrings of sentences, whether they consist of one word, or more than one. Finally the strings in the covering lexicon may contain overlapping words even though those in the domain of an individual interpretation function must not, since those overlapping words could have occurred in different sentences.

Finally, we are ready to define the learning problem at hand.

**The Lexicon Acquisition Problem:**
**Given:** a multiset of strings $S = \{s_1, \ldots, s_n\}$ and a multiset of labeled trees $T = \{t_1, \ldots, t_n\}$,
**Find:** a multiset of interpretation functions, $F = \{f_1, \ldots, f_n\}$, such that the cardinality of the covering lexicon expressed by $STF = \{< s_1, t_1, f_1 >, \ldots, < s_n, t_n, f_n >\}$ is minimized. If such a set is found, we say we have found a *minimal* set of interpretations (or a *minimal covering lexicon*). □

Less formally, a learner is presented with a multiset of sentences ($S$) paired with their meanings ($T$); the goal of learning is to find the smallest lexicon consistent with this data. This lexicon is the paired listing of all phrases occurring in the domain of some $f_i \in F$ (where $F$ is the multiset of interpretation functions found) with each of the elements in their meaning sets. The motivation for finding a lexicon of minimal size is the usual bias towards simplicity of representation and generalization beyond the training data. While this definition allows for phrases of any length, we will usually want to limit the length of phrases to be considered for inclusion in the domain of the interpretation functions, for efficiency purposes.

Once we determine a set of interpretation functions for a set of strings and trees, there is only one unique covering lexicon expressed by $STF$. However, this might not be the only set of interpretation functions possible, and may not result in the lexicon with smallest cardinality. For example, the covering lexicon given with the previous example is not a *minimal* covering lexicon. For the two sentences given, we could find minimal, though rather degenerate, lexicons such as:

```
{   ("girl",    [ingest, agent:[person, sex:female, age:child],
                  patient:[food, type:pasta, accomp:[food, type:cheese]]]),
    ("man",     [ingest, agent:[person, sex:male, age:adult],
                  patient:[food, type:cheese]]) }
```

This type of lexicon becomes less likely as the size of the corpus grows.

## 3.2 Implications of the Definition

This definition of the lexicon acquisition problem differs from that given by other authors, including Riloff and Jones (1999), Siskind (1996), Manning (1993), Brent (1991) and others, as further discussed in Section 7. Our definition of the problem makes some assumptions about the training input. First, by making $f$ a function instead of a relation, the definition assumes that the meaning for each phrase in a sentence appears once in the representation of that sentence, the *single-use* assumption. Second, by making $f$ one-to-one, it assumes *exclusivity*, that each vertex in a sentence's representation is due to only one phrase in the sentence. Third, it assumes that a phrase's meaning is a connected subgraph of a sentence's representation, not a more distributed representation, the *connectedness* assumption. While the first assumption may not hold for some representation languages, it does not present a problem in the domains we have considered. The second and third assumptions are perhaps less problematic with respect to general language use.

Our definition also assumes *compositionality*: that the meaning of a sentence is derived from the meanings of the phrases it contains, in addition, perhaps to some "connecting" information specific to the representation at hand, but is not derived from external sources such as noise. In other words, all the vertices of a sentence's representation are included within the meaning of some word or phrase in that sentence. This assumption is similar to the linking rules of Jackendoff (1990), and has been used in previous work on grammar and language acquisition (e.g., Haas and Jayaraman, 1997; Siskind, 1996[4]) While there is some debate in the linguistics community about the ability of compositional techniques to handle all phenomena (Fillmore, 1988; Goldberg, 1995), making this assumption simplifies the learning process and works reasonably for the domains of interest here. Also, since we allow multi-word phrases in the lexicon (e.g., ("kick the bucket", `die(_)`)), one objection to compositionality can be addressed.

This definition also allows training input in which:

1. Words and phrases have multiple meanings. That is, homonymy might occur in the lexicon.

2. Several phrases map to the same meaning. That is, synonymy might occur in the lexicon.

3. Some words in a sentence do not map to any meanings, leaving them unused in the assignment of words to meanings.[5]

4. Phrases of contiguous words map to parts of a sentence's meaning representation.

Of particular note is lexical ambiguity (1 above). Note that we could have also derived an ambiguous lexicon such as:

> { ("girl", `[person, sex:female, age:child]`),
> ("ate", `[ingest]`),
> ("ate", `[ingest, agent:[person, sex:male, age:adult]]`),
> ("pasta", `[food, type:pasta]`),
> ("the cheese", `[food, type:cheese]`) }.

from our sample corpus. In this lexicon, "ate" is an ambiguous word. The earlier example minimizes ambiguity resulting in an alternative, more intuitively pleasing lexicon. While our problem definition first minimizes the number of entries in the lexicon, our learning algorithm will also exploit a preference for minimizing ambiguity.

Also note that our definition allows training input in which sentences themselves are ambiguous (paired with more than one meaning), since a given sentence in $S$ (a multiset) might appear multiple times appear with more than one meaning. In fact, the training data that we consider in Section 5 does have some ambiguous sentences.

Our definition of the lexicon acquisition problem does not fit cleanly into the traditional definition of learning for classification. Each training example contains a sentence and its semantic parse, and we are trying to extract semantic information about some of the phrases in that sentence. So each example potentially contains information about multiple target concepts (phrases), and we are trying to pick out the relevant "features," or vertices of the representation, corresponding to the correct meaning of each phrase. Of course, our assumptions of single-use, exclusivity, connectedness, and compositionality impose additional constraints. In addition to this "multiple examples in one" learning scenario, we do not have access to negative examples, nor can we derive any implicit negatives, because of the possibility of ambiguous and synonymous phrases.

---

4. In fact, all of these assumptions except for single-use were made by Siskind (1996); see Section 7 for details.

5. These words may, however, serve as cues to a parser on how to assemble sentence meanings from word meanings.

In some ways the problem is related to clustering, which is also capable of learning multiple, potentially non-disjoint categories. However, it is not clear how a clustering system could be made to learn the phrase-meaning mappings needed for parsing. Finally, current systems that learn multiple concepts commonly use examples for other concepts as negative examples of the concept currently being learned. The implicit assumption made by doing this is that concepts are disjoint, an unwarranted assumption in the presence of synonymy.

## 4. The WOLFIE Algorithm and an Example

In this section, we first discuss some issues we considered in the design of our algorithm, then describe it fully in Section 4.2.

### 4.1 Solving the Lexicon Acquisition Problem

A first attempt to solve the Lexicon Acquisition Problem might be to examine all interpretation functions across the corpus, then choose the one(s) with minimal lexicon size. The number of possible interpretation functions for a given input pair is dependent on both the size of the sentence and its representation. In a sentence with $w$ words, there are $\Theta(w^2)$ possible phrases, not a particular challenge.

However, the number of possible interpretation functions grows extremely quickly with the size of the input. For a sentence with $p$ phrases and an associated tree with $n$ vertices, the number of possible interpretation functions is:

$$c!(n-1)! \sum_{i=1}^{c} \frac{1}{(i-1)!(n-i)!(c-i)!}. \tag{1}$$

where $c$ is $min(p, n)$. The derivation of the above formula is as follows. We must choose which phrases to use in the domain of $f$, and we can choose one phrase, or two, or any number up to $min(p, n)$ (if $n < p$ we can only assign $n$ phrases since $f$ is one-to-one), or

$$\left( \begin{array}{c} p \\ i \end{array} \right) = \frac{p!}{i!(p-i)!}$$

where $i$ is the number of phrases chosen. But we can also permute these phrases, so that the "order" in which they are assigned to the vertices is different. There are $i!$ such permutations. We must also choose which vertices to include in the range of the interpretation function. We have to choose the root each time, so if we are choosing $i$ vertices, we have $n-1$ choose $i-1$ vertices left after choosing the root, or

$$\left( \begin{array}{c} n-1 \\ i-1 \end{array} \right) = \frac{(n-1)!}{(i-1)!(n-i)!}.$$

The full number of possible interpretation functions is then:

$$\sum_{i=1}^{min(p,n)} \frac{p!}{i!(p-i)!} \times i! \times \frac{(n-1)!}{(i-1)!(n-i)!},$$

which simplifies to Equation 1. When $n = p$, the largest term of this equation is $c! = p!$, which grows at least exponentially with $p$, so in general the number of interpretation functions is too large to allow enumeration. Therefore, finding a lexicon by examining all interpretations across the corpus, then choosing the lexicon(s) of minimum size, is clearly not tractable.

Instead of finding all interpretations, one could find a set of candidate meanings for each phrase, from which the final meaning(s) for that phrase could be chosen in a way that minimizes lexicon size. One way to find candidate meanings is to *fracture* the meanings of sentences in which a phrase appears. Siskind (1993) defined fracturing (he also calls it the Unlink* operation) over terms such that the result includes all subterms of an expression plus ⊥. In our representation formalism, this corresponds to finding all possible connected subgraphs of a meaning, and adding the empty graph. Like the interpretation function technique just discussed, fracturing would also lead to an exponential blowup in the number of candidate meanings for a phrase: A lower bound on the number of connected subgraphs for a full binary tree with $n$ vertices is obtained by noting that any subset of the $(n + 1)/2$ leaves may be deleted and still maintain connectivity of the remaining tree. Thus, counting all of the ways that leaves can be deleted gives us a lower bound of $2^{(n+1)/2}$ fractures.[6] This does not completely rule out fracturing as part of a technique for lexicon learning since trees do not tend to get very large, and indeed Siskind uses it in many of his systems, with other constraints to help control the search. However, we wish to avoid any chance of exponential blowup to preserve the generality of our approach for other tasks.

Another option is to force Chill to essentially induce a lexicon on its own. In this model, we would provide to Chill an ambiguous lexicon in which each phrase is paired with every fracture of every sentence in which it appears. Chill would then have to decide which set of fractures leads to the correct parse for each training sentence, and would only include those in a final learned parser-lexicon combination. Thus the search would again become exponential. Furthermore, even with small representations, it would likely lead to a system with poor generalization ability. While some of Siskind's work (e.g., Siskind, 1992) took syntactic constraints into account and did not encounter such difficulties, those versions did not handle lexical ambiguity.

If we could efficiently find some good candidates, a standard induction algorithm could then attempt to use them as a source of training examples for each phrase. However, any attempt to use the list of candidate meanings of one phrase as negative examples for another phrase would be flawed. The learner could not know in advance which phrases are possibly synonymous, and thus which phrase lists to use as negative examples of other phrase meanings. Also, many representation components would be present in the lists of more than one phrase. This is a source of conflicting evidence for a learner, even without the presence of synonymy. Since only positive examples are available, one might think of using most specific conjunctive learning, or finding the intersection of all the representations

---

6. Thanks to net-citizen Dan Hirshberg for help with this analysis.

---

**For** each phrase, $p$ (of at most two words):
   1.1) Collect the training examples in which $p$ appears
   1.2) Calculate LICS from (sampled) pairs of these examples' representations
   1.3) For each $l$ in the LICS, add $(p, l)$ to the set of candidate lexicon entries
**Until** the input representations are covered, or no candidate lexicon entries remain **do**:
   2.1) Add the best (phrase, meaning) pair from the candidate entries to the lexicon
   2.2) Update candidate meanings of phrases in the same sentences as the phrase just learned
**Return** the lexicon of learned (phrase, meaning) pairs.

---

Figure 6: WOLFIE Algorithm Overview

for each phrase, as proposed by Anderson (1977). However, the meanings of an ambiguous phrase are disjunctive, and this intersection would be empty. A similar difficulty would be expected with the positive-only compression of Muggleton (1995).

## 4.2 Our Solution: WOLFIE

The above analysis leads us to believe that the Lexicon Acquisition Problem is computationally intractable. Therefore, we can not perform an efficient search for the best lexicon. Nor can we use a standard induction algorithm. Therefore, we have implemented WOLFIE[7], outlined in Figure 6, which finds an approximate solution to the Lexicon Acquisition Problem. Our approach is to generate a set of candidate lexicon entries, from which the final learned lexicon is derived by greedily choosing the "best" lexicon item at each point, in the hopes of finding a final (minimal) covering lexicon. We do not actually learn interpretation functions, so do not guarantee that we will find a covering lexicon.[8] Even if we were to search for interpretation functions, using a greedy search would also not guarantee covering the input, and of course it also does not guarantee that a minimal lexicon is found. However, we will later present experimental results demonstrating that our greedy approach performs well.

WOLFIE first derives an initial set of candidate meanings for each phrase. The algorithm for generating candidates, LICS, attempts to find a "maximally common" meaning for each phrase, which biases toward both finding a small lexicon by covering many vertices of a tree at once, and finding a lexicon that actually does cover the input. Second, WOLFIE chooses final lexicon entries from this candidate set, one at a time, updating the candidate set as it goes, taking into account our assumptions of single-use, connectedness, and exclusivity. The basic scheme for choosing entries from the candidate set is to maximize the prediction of meanings given phrases, but also to find general meanings. This adds a tension between LICS, which cover many vertices, and generality, which biases towards fewer vertices. However, generality, like LICS, helps lead to a small lexicon since a general meaning will more likely apply widely across a corpus.

---

7. The code is available upon request from the first author.

8. Though, of course, interpretation functions are not the only way to guarantee a covering lexicon – see Siskind (1993) for an alternative.

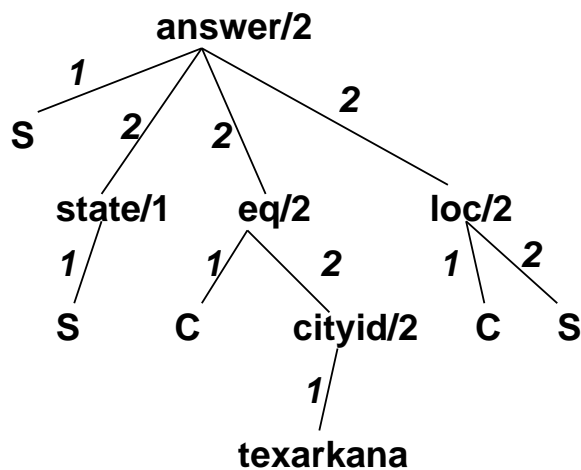

Figure 7: Tree with Variables

Let us explain the algorithm in further detail by way of an example, using Spanish instead of English to illustrate the difficulty somewhat more clearly. Consider the following corpus:

1. ¿ Cuál es el capital del estado con la población más grande?
   `answer(C, (capital(S,C), largest(P, (state(S), population(S,P)))))`.

2. ¿ Cuál es la punta más alta del estado con la area más grande?
   `answer(P, (high_point(S,P), largest(A, (state(S), area(S,A)))))`.

3. ¿ En que estado se encuentra Texarkana?
   `answer(S, (state(S), eq(C,cityid(texarkana,_)), loc(C,S)))`.

4. ¿ Qué capital es la más grande?
   `answer(A, largest(A, capital(A)))`.

5. ¿ Qué es la area de los estados unitos?
   `answer(A, (area(C,A), eq(C,countryid(usa))))`.

6. ¿ Cuál es la población de un estado que bordean a Utah?
   `answer(P, (population(S,P), state(S), next_to(S,M), eq(M,stateid(utah))))`.

7. ¿ Qué es la punta más alta del estado con la capital Madison?
   `answer(C, (high_point(B,C), loc(C,B), state(B),`
   `    capital(B,A), eq(A,cityid(madison,_))))`.

The sentence representations here are slightly different than the tree representations given in the problem definition, with the main difference being the addition of existentially quantified variables shared between some leaves of a representation tree. As mentioned in Section 2.1, the representations are Prolog queries to a database. Given such a query, we can create a tree that conforms to our formalism, but with this addition of quantified variables. An example is shown in Figure 7 for the representation of the third sentence. Each vertex is a predicate name and its arity, in the Prolog style, e.g., `state/1`, with quantified variables at some of the leaves. For each outgoing edge $(n, m)$ of a vertex $n$, the edge is labeled with the argument position filled by the subtree rooted by $m$. If there is not an edge labeled with a given argument position, the argument is a free variable. Each vertex labeled with a variable (which can occur only at leaves) is an existentially quantified variable whose scope is the entire tree (or query). The learned lexicon, however, does not need to maintain the identity between variables across distinct lexical entries.

Another representation difference is that we will strip the `answer` predicate from the input to our learner,[9] thus allowing a forest of directed trees as input rather than a single tree. The definition of the problem easily extends such that the root of each tree in the forest must be in the domain of some interpretation function.

Evaluation of our system using this representation is given in Section 5.1; evaluation using a representation without variables or forests is presented in Section 5.2. We previously (Thompson, 1995) presented results demonstrating learning representations of a different form, that of a case-role representation (Fillmore, 1968) augmented with Conceptual Dependency (Schank, 1975) information. This last representation conforms directly to our problem definition.

Now, continuing with the example of solving the Lexicon Acquisition Problem for this corpus, let us also assume for simplification, although not required, that sentences are stripped of phrases that we know have empty meanings (e.g., "qué", "es", "con", and "la"). We will similarly assume that it is known that some phrases refer directly to given database constants (e.g., location names), and remove those phrases and their meaning from the training input.

### 4.2.1 CANDIDATE GENERATION PHASE

Initial candidate meanings for a phrase are produced by computing the maximally common substructure(s) between sampled pairs of representations of sentences that contain it. We derive common substructure by computing the Largest Isomorphic Connected Subgraphs (LICS) of two labeled trees, taking labels into account in the isomorphism. The analogous Largest Common Subgraph problem (Garey & Johnson, 1979) is solvable in polynomial time if, as we assume, both inputs are trees and if $K$, the number of edges to include, is given. Thus, we start with $K$ set equal to the largest number of edges in the two trees being compared, test for common subgraph(s), and iterate down to $K = 1$, stopping when one or more subgraphs are found for a given $K$.

For the Prolog query representation, the algorithm is complicated a bit by variables. Therefore, we use LICS with an addition similar to computing the Least General Generalization of first-order clauses (Plotkin, 1970). The LGG of two sets of literals is the least general set of literals that subsumes both sets of literals. We add to this by allowing that when a term in the argument of a literal is a conjunction, the algorithm tries all orderings in its matching of the terms in the conjunction. Overall, our algorithm for finding the LICS between two trees in the Prolog representation first finds the common labeled edges and vertices as usual in LICS, but treats all variables as equivalent. Then, it computes the Least General Generalization, with conjunction taken into account, of the resulting trees as converted back into literals. For example, given the two trees:

---

9. The predicate is omitted because CHILL initializes the parse stack with the `answer` predicate, and thus no word has to be mapped to it.

| Phrase | LICS | From Sentences |
|--------|------|----------------|
| "capital": | `largest(_,_)` | 1,4 |
|  | `capital(_,_)` | 1,7 |
|  | `state(_)` | 1,7 |
| "grande": | `largest(_,state(_))` | 1,2 |
|  | `largest(_,_)` | 1,4; 2,4 |
| "estado": | `largest(_,state(_))` | 1,2 |
|  | `state(_)` | 1,3; 1,7; 2,3; 2,6; 2,7; 3,6; 6,7 |
|  | `(population(S,_), state(S))` | 1,6 |
|  | `capital(_,_)` | 1,7 |
|  | `high_point(_,_)` | 2,7 |
|  | `(state(S), loc(_,S))` | 3,7 |
| "punta mas": | `high_point(_,_)` | 2,7 |
|  | `state(_)` | 2,7 |
| "encuentra": | `(state(S), loc(_,S))` | 3 |

Table 1: Sample Candidate Lexical Entries and their Derivation

```
answer(C, (largest(P, (state(S), population(S,P))), capital(S,C))).

answer(P, (high_point(S,P), largest(A, (state(S), area(S,A)))))).,
```

the common meaning is `answer(_,largest(_,state(_))`. Note that the LICS of two trees may not be unique: there may be multiple common subtrees that both contain the same number of edges; in this case LICS returns multiple answers.

The sets of initial candidate meanings for some of the phrases in the sample corpus are shown in Table 1. While in this example we show the LICS for all pairs that a phrase appears in, in the actual algorithm we randomly sample a subset for efficiency reasons, as in GOLEM (Muggleton & Feng, 1990). For phrases appearing in only one sentence (e.g., "encuentra"), the entire sentence representation (excluding the database constant given as background knowledge) is used as an initial candidate meaning. Such candidates are typically generalized in step 2.2 of the algorithm to only the correct portion of the representation before they are added to the lexicon; we will see an example of this below.

### 4.2.2 Adding to the Final Lexicon

After deriving initial candidates, the greedy search begins. The heuristic used to evaluate candidates attempts to help assure that a small but covering lexicon is learned. The heuristic first looks at the weighted sum of two components, where $p$ is the phrase and $m$ its candidate meaning:

1. $P(m \mid p) \times P(p \mid m) \times P(m) = P(p) \times P(m \mid p)^2$

2. The generality of $m$

Then, ties in this value are broken by preferring less ambiguous (those with fewer current meanings) and shorter phrases. The first component is analogous the cluster evaluation heuristic used by Cobweb (Fisher, 1987), which measures the utility of clusters based on attribute-value pairs and categories, instead of meanings and phrases. The probabilities are estimated from the training data and then updated as learning progresses to account for phrases and meanings already covered. We will see how this updating works as we continue through our example of the algorithm. The goal of this part of the heuristic is to maximize the probability of predicting the correct meaning for a randomly sampled phrase. The equality holds by Bayes Theorem. Looking at the right side, $P(m \mid p)^2$ is the expected probability that meaning $m$ is correctly guessed for a given phrase, $p$. This assumes a strategy of probability matching, in which a meaning $m$ is chosen for $p$ with probability $P(m \mid p)$ and correct with the same probability. The other term, $P(p)$, biases the component by how common the phrase is. Interpreting the left side of the equation, the first term biases towards lexicons with low ambiguity, the second towards low synonymy, and the third towards frequent meanings.

The second component of the heuristic, *generality*, is computed as the negation of the number of vertices in the meaning's tree structure, and helps prefer smaller, more general meanings. For example, in the candidate set above, if all else were equal, the generality portion of the heuristic would prefer `state(_)`, with generality value -1, over `largest(_,state(_))` and `(state(S),loc(_,S))`, each with generality value -2, as the meaning of "estado". Learning a meaning with fewer terms helps evenly distribute the vertices in a sentence's representation among the meanings of the phrases in that sentence, and thus leads to a lexicon that is more likely to be correct. To see this, we note that some pairs of words tend to frequently co-occur ("grande" and "estado" in our example), and so their joint representation (meaning) is likely to be in the set of candidate meanings for both words. By preferring a more general meaning, we easily ignore these incorrect joint meanings.

In this example and all experiments, we use a weight of 10 for the first component of the heuristic, and a weight of 1 for the second. The first component has smaller absolute values and is therefore given a higher weight. Modulo this consideration, results are not overly-sensitive to the weights and automatically setting them using cross-validation on the training set (Kohavi & John, 1995) had little effect on overall performance. In Table 2 we illustrate the calculation of the heuristic measure for some of the above fourteen pairs, and its value for all. The calculation shows the sum of multiplying 10 by the first component of the heuristic and multiplying 1 by the second component. The first component is simplified as follows:

$$P(p) \times P(m \mid p)^2 = \frac{\mid p \mid}{t} \times \frac{\mid m \cap p \mid^2}{\mid p \mid^2} \approx \frac{\mid m \cap p \mid^2}{\mid p \mid},$$

where $\mid p \mid$ is the number of times phrase $p$ appears in the corpus, $t$ is the initial number of candidate phrases, and $\mid m \cap p \mid$ is the number of times that meaning $m$ is paired with phrase $p$. We can ignore $t$ since the number of phrases in the corpus is the same for each pair, and has no effect on the ranking. The highest scoring pair is ("estado", `state(_)`), so it is added to the lexicon.

Next is the candidate generalization step (2.2), described algorithmically in Figure 8. One of the key ideas of the algorithm is that each phrase-meaning choice can constrain the candidate meanings of phrases yet to be learned. Given the assumption that each portion of the representation is due to at most one phrase in the sentence (exclusivity), once part of a

| Candidate Lexicon Entry | Heuristic Value |
|---|---|
| ("capital", `largest(_,_)`): | $10(2^2/3) + 1(-1) = 12.33$ |
| ("capital", `capital(_,_)`): | 12.33 |
| ("capital", `state(_,_)`): | 12.33 |
| ("grande", `largest(_,state(_))`): | $10(2^2/3) + 1(-2) = 11.3$ |
| ("grande", `largest(_,_)`): | 29 |
| ("estado", `largest(_,state(_))`): | $10(2^2/5) + 1(-2) = 6$ |
| ("estado", `state(_)`): | $10(5^2/5) + 1(-1) = 49$ |
| ("estado", `(population(S,_), state(S))`): | 6 |
| ("estado", `capital(_,_)`): | 7 |
| ("estado", `high_point(_,_)`): | 7 |
| ("estado", `(state(S), loc(_,S))`): | 6 |
| ("punta mas", `high_point(_,_)`): | 19 |
| ("punta mas", `state(_)`): | $10(2^2/2) + 1(-1) = 19$ |
| ("encuentra", `(state(S), loc(_,S))`): | $10(1^2/1) + 1(-2) = 8$ |

Table 2: Heuristic Value of Sample Candidate Lexical Entries

---

**Given:** A learned phrase-meaning pair $(l, g)$

**For** all sentence-representation pairs containing $l$ and $g$, mark them as covered.
**For** each candidate phrase-meaning pair $(p, m)$:
    **If** $p$ occurs in some training pairs with $(l, g)$ **then**
        **If** the vertices of $m$ intersect the vertices of $g$ **then**
            **If** all occurrences of $m$ are now covered **then**
                Remove $(p, m)$ from the set of candidate pairs.
            **Else**
                Adjust the heuristic value of $(p, m)$ as needed to account
                    for newly covered nodes of the training representations.
            Generalize $m$ to remove covered nodes, obtaining $m'$, and
            Calculate the heuristic value of the new candidate pair $(p, m')$.
**If** no candidate meanings remain for an uncovered phrase **then**
    Derive new LICS from uncovered representations and
        calculate their heuristic values.

---

Figure 8: The Candidate Generalization Phase

representation is covered, no other phrase in the sentence can be paired with that meaning (at least for that sentence). Therefore, in step 2.2 the candidate meanings for words in the same sentences as the word just learned are generalized to exclude the representation just learned. We use an operation analogous to set difference when finding the remaining uncovered vertices of the representation when generalizing meanings to eliminate covered vertices from candidate pairs. For example, if the meaning `largest(_,_)` were learned for a phrase in sentence 2, the meaning left behind would be a forest consisting of the trees `high_point(S,_)` and `(state(S), area(S,_))`. Also, if the generalization results in an empty tree, new LICS are calculated. In our example, since `state(_)` is covered in sentences 1, 2, 3, 6, and 7, the candidates for several other words in those sentences are generalized. For example, the meaning `(state(S), loc(_,S))` for "encuentra", is generalized to `loc(_,_)`, with a new heuristic value of $10(1^2/1) + 1(-1) = 9$. Also, our single-use assumption allows us to remove all candidate pairs containing "estado" from the set of candidate meanings, since the learned pair covers all occurrences of "estado" in that set.

Note that the pairwise matchings to generate candidate items, together with this updating of the candidate set, enable multiple meanings to be learned for ambiguous phrases, and makes the algorithm less sensitive to the initial rate of sampling for LICS. For example, note that "capital" is ambiguous in this data set, though its ambiguity is an artifact of the way that the query language was designed, and one does not ordinarily think of it as an ambiguous word. However, both meanings will be learned: The second pair added to the final lexicon is ("grande", `largest(_,_)`), which causes a generalization to the empty meaning for the first candidate entry in Table 2, and since no new LICS from sentence 4 can be generated, its entire remaining meaning is added to the candidate meaning set for both "capital" and "más."

Subsequently, the greedy search continues until the resulting lexicon covers the training corpus, or until no candidate phrase meanings remain. In rare cases, learning errors occur that leave some portions of representations uncovered. In our example, the following lexicon is learned:

("estado", `state(_)`),

("grande", `largest(_)`),

("area", `area(_)`),

("punta", `high_point(_,_)`),

("población", `population(_,_)`),

("capital", `capital(_,_)`),

("encuentra", `loc(_,_)`),

("alta", `loc(_,_)`),

("bordean", `next_to(_)`),

("capital", `capital(_)`).

In the next section, we discuss the ability of WOLFIE to learn lexicons that are useful for parsers and parser acquisition.

## 5. Evaluation of Wolfie

The following two sections discuss experiments testing Wolfie's success in learning lexicons for both real and artificial corpora, comparing it in several cases to a previously developed lexicon learning system.

### 5.1 A Database Query Application

This section describes our experimental results on a database query application. The first corpus discussed contains 250 questions about U.S. geography, paired with their Prolog query to extract the answer to the question from a database. This domain was originally chosen due to the availability of a hand-built natural language interface, Geobase, to a database containing about 800 facts. Geobase was supplied with Turbo Prolog 2.0 (Borland International, 1988), and designed specifically for this domain. The questions in the corpus were collected by asking undergraduate students to generate English questions for this database, though they were given only cursory knowledge of the database without being given a chance to use it. To broaden the test, we had the same 250 sentences translated into Spanish, Turkish, and Japanese. The Japanese translations are in word-segmented Roman orthography. Translated questions were paired with the appropriate logical queries from the English corpus.

To evaluate the learned lexicons, we measured their utility as background knowledge for Chill. This is performed by choosing a random set of 25 test examples and then learning lexicons and parsers from increasingly larger subsets of the remaining 225 examples (increasing by 50 examples each time). After training, the test examples are parsed using the learned parser. We then submit the resulting queries to the database, compare the answers to those generated by submitting the correct representation to the database, and record the percentage of correct (matching) answers. By using the difficult "gold standard" of retrieving a correct answer, we avoid measures of partial accuracy that we believe do not adequately measure final utility. We repeated this process for ten different random training and test sets and evaluated performance differences using a two-tailed, paired $t$-test with a significance level of $p \leq 0.05$.

We compared our system to an incremental (on-line) lexicon learner developed by Siskind (1996). To make a more equitable comparison to our batch algorithm, we ran his in a "simulated" batch mode, by repeatedly presenting the corpus 500 times, analogous to running 500 epochs to train a neural network. While this does not actually add new kinds of data over which to learn, it allows his algorithm to perform inter-sentential inference in both directions over the corpus instead of just one. Our point here is to compare accuracy over the same size training corpus, a metric not optimized for by Siskind. We are not worried about the difference in execution time here,[10] and the lexicons learned when running Siskind's system in incremental mode (presenting the corpus a single time) resulted in substantially lower performance in preliminary experiments with this data. We also removed Wolfie's ability to learn phrases of more than one word, since the current version of Siskind's system does not have this ability. Finally, we made comparisons to the parsers learned by CHILL when using a hand-coded lexicon as background knowledge.

---

10. The CPU times of the two system are not directly comparable since one is written in Prolog and the other in Lisp. However, the learning time of the two systems is approximately the same if Siskind's system is run in incremental mode, just a few seconds with 225 training examples.

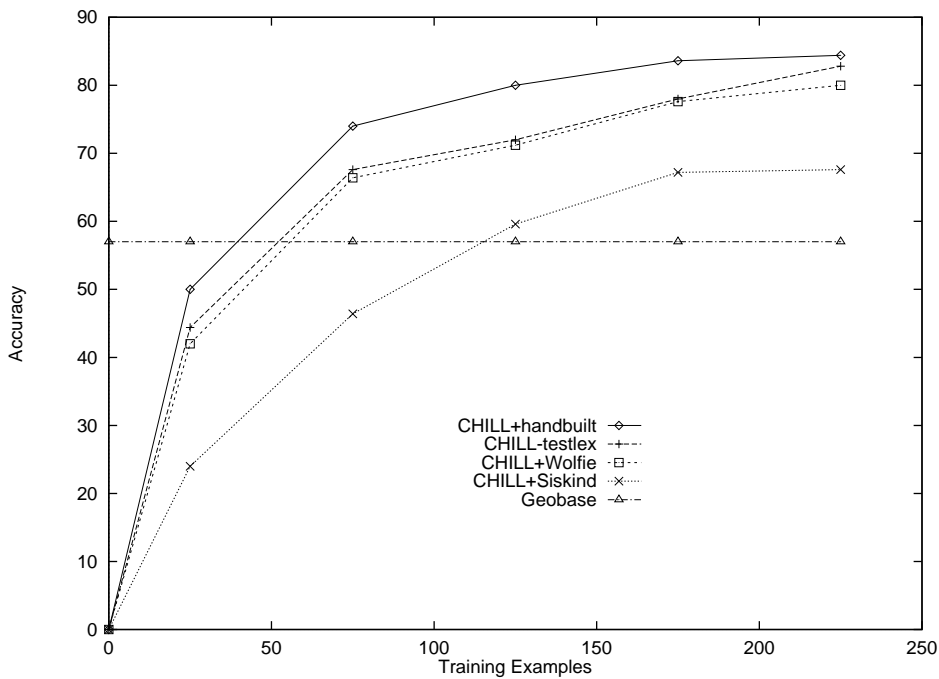

Figure 9: Accuracy on English Geography Corpus

In this and similar applications, there are many terms, such as state and city names, whose meanings can be automatically extracted from the database. Therefore, all tests below were run with such names given to the learner as an initial lexicon; this is helpful but not required. Section 5.2 gives results for a different task with no such initial lexicon. However, unless otherwise noted, for all tests within this Section (5.1) we did not strip sentences of phrases known to have empty meanings, unlike in the example of Section 4.

### 5.1.1 Comparisons using English

The first experiment was a comparison on the original English corpus. Figure 9 shows learning curves for CHILL when using the lexicons learned by WOLFIE (CHILL+Wolfie) and by Siskind's system (CHILL+Siskind). The uppermost curve (CHILL+handbuilt) shows CHILL's performance when given the hand-built lexicon. CHILL-testlex shows the performance when words that never appear in the training data (e.g., are only in the test sentences) are deleted from the hand-built lexicon (since a learning algorithm has no chance of learning these). Finally, the horizontal line shows the performance of the GEOBASE benchmark.

The results show that a lexicon learned by WOLFIE led to parsers that were almost as accurate as those generated using a hand-built lexicon. The best accuracy is achieved by parsers using the hand-built lexicon, followed by the hand-built lexicon with words only in the test set removed, followed by WOLFIE, followed by Siskind's system. All the systems do as well or better than GEOBASE by the time they reach 125 training examples. The differences between WOLFIE and Siskind's system are statistically significant at all training example sizes. These results show that Wolfie can learn lexicons that support the learning of successful parsers, and that are better from this perspective than those learned by a competing system. Also, comparing to the `CHILL-testlex` curve, we see that most of the drop in accuracy from a hand-built lexicon is due to words in the test set that the system has not seen during training. In fact, none of the differences between `CHILL+Wolfie` and `CHILL-testlex` are statistically significant.

| Lexicon | Coverage | Ambiguity | Entries |
|---------|----------|-----------|---------|
| hand-built | 100% | 1.2 | 88 |
| Wolfie | 100% | 1.1 | 56.5 |
| Siskind | 94.4% | 1.7 | 154.8 |

Table 3: Lexicon Comparison

One of the implicit hypotheses of our problem definition is that coverage of the training data implies a good lexicon. The results show a coverage of 100% of the 225 training examples for Wolfie versus 94.4% for Siskind. In addition, the lexicons learned by Siskind's system were more ambiguous and larger than those learned by Wolfie. Wolfie's lexicons had an average of 1.1 meanings per word, and an average size of 56.5 entries (after 225 training examples) versus 1.7 meanings per word and 154.8 entries in Siskind's lexicons. For comparison, the hand-built lexicon had 1.2 meanings per word and 88 entries. These differences, summarized in Table 3, undoubtedly contribute to the final performance differences.

### 5.1.2 Performance for Other Natural Languages

Next, we examined the performance of the two systems on the Spanish version of the corpus. Figure 10 shows the results. The differences between using Wolfie and Siskind's learned lexicons for Chill are again statistically significant at all training set sizes. We also again show the performance with hand-built lexicons, both with and without phrases present only in the testing set. The performance compared to the hand-built lexicon with test-set phrases removed is still competitive, with the difference being significant only at 225 examples.

Figure 11 shows the accuracy of learned parsers with Wolfie's learned lexicons for all four languages. The performance differences among the four languages are quite small, demonstrating that our methods are not language dependent.

### 5.1.3 A Larger Corpus

Next, we present results on a larger, more diverse corpus from the geography domain, where the additional sentences were collected from computer science undergraduates in an introductory AI course. The set of questions in the smaller corpus was collected from students in a German class, with no special instructions on the complexity of queries desired. The AI students tended to ask more complex and diverse queries: their task was to give five interesting questions and the associated logical form for a homework assignment, though again they did not have direct access to the database. They were requested to give at least one sentence whose representation included a predicate containing embedded predicates, for example `largest(S, state(S))`, and we asked for variety in their sentences. There were 221 new sentences, for a total of 471 (including the original 250 sentences).

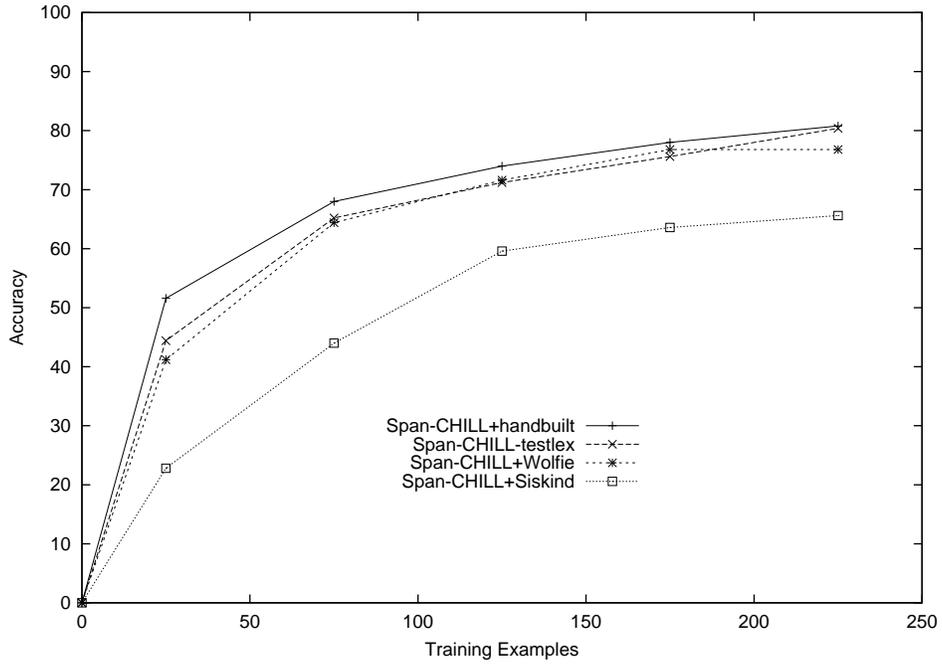

Figure 10: Accuracy on Spanish

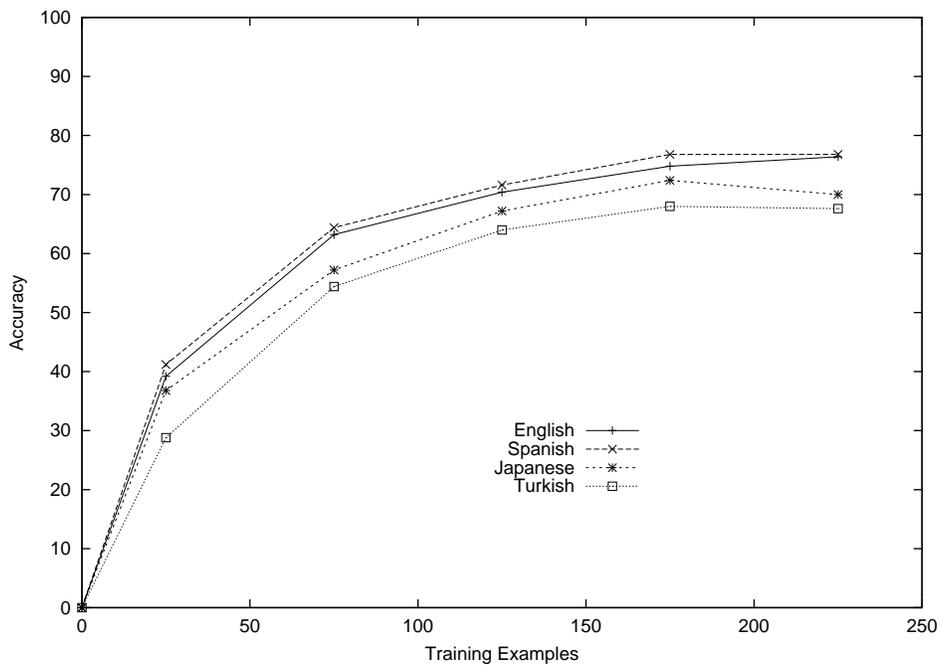

Figure 11: Accuracy on All Four Languages

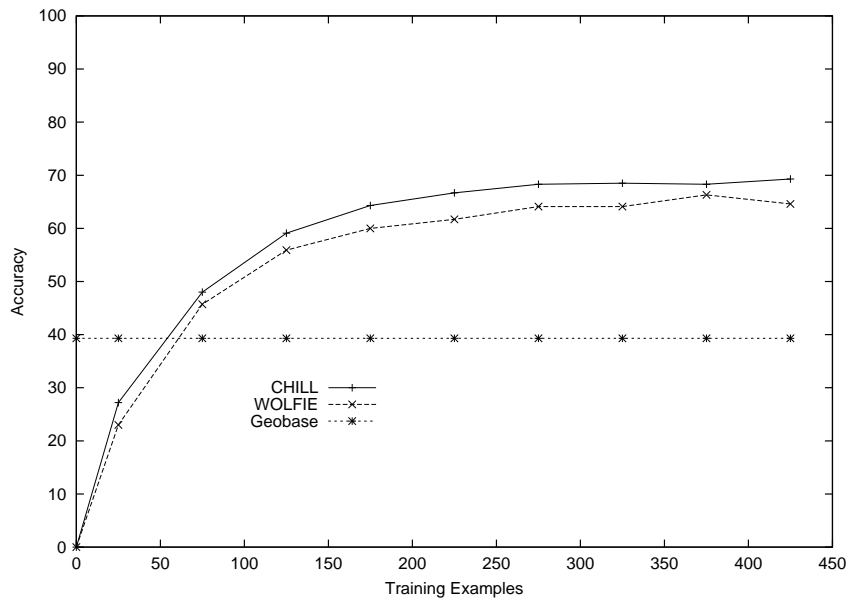

Figure 12: Accuracy on the Larger Geography Corpus

For these experiments, we split the data into 425 training sentences and 46 test sentences, for 10 random splits, then trained Wolfie and then Chill as before. Our goal was to see whether Wolfie was still effective for this more difficult corpus, since there were approximately 40 novel words in the new sentences. Therefore, we tested against the performance of Chill with an extended hand-built lexicon. For this test, we stripped sentences of phrases known to have empty meanings, as in the example of Section 4.2. Again, we did not use phrases of more than one word, since these do not seem to make a significant difference in this domain. For these results, we compare Wolfie's lexicons for Chill using hand-built lexicons without phrases that only appear in the test set.

Figure 12 shows the resulting learning curves. The differences between Chill using the hand-built and learned lexicons are statistically significant at 175, 225, 325, and 425 examples (four out of the nine data points). The more mixed results here indicate both the difficulty of the domain and the more variable vocabulary. However, the improvement of machine learning methods over the Geobase hand-built interface is much more dramatic for this corpus.

### 5.1.4 LICS versus Fracturing

One component of the algorithm not yet evaluated explicitly is the candidate generation method. As mentioned in Section 4.1, we could use fractures of representations of sentences in which a phrase appears to generate the candidate meanings for that phrase, instead of LICS. We used this approach and compared it to the previously described method of using the largest isomorphic connected subgraphs of sampled pairs of representations as candidate meanings. To attempt a more fair comparison, we also sampled representations for fracturing, using the same number of source representations as the number of pairs sampled for LICS.

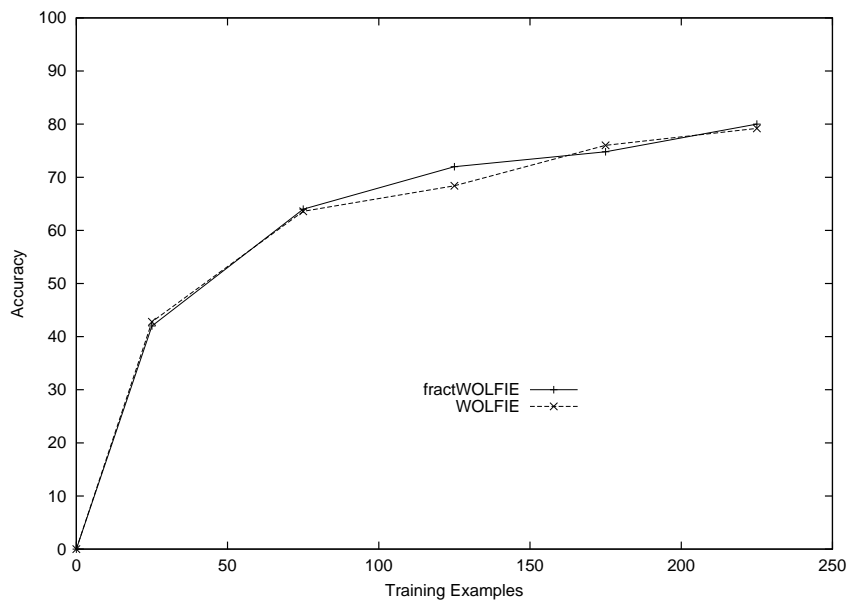

Figure 13: Fracturing vs. LICS: Accuracy

The accuracy of CHILL when using the resulting learned lexicons as background knowledge are shown in Figure 13. Using fracturing (`fractWOLFIE`) shows little or no advantage; none of the differences between the two systems are statistically significant.

In addition, the number of initial candidate lexicon entries from which to choose is much larger for fracturing than our LICS method, as shown in Figure 14. This is true even though we sampled the same number of representations as pairs for LICS, because there are a larger number of fractures for an arbitrary representation than the number of LICS for an arbitrary pair. Finally, WOLFIE's learning time when using fracturing is greater than that when using LICS, as shown in Figure 15, where the CPU time is shown in seconds.

In summary, these differences show the utility of LICS as a method for generating candidates: a more thorough method does not result in better performance, and also results in longer learning times. One could claim that we are handicapping fracturing since we are only sampling representations for fracturing. This may indeed help the accuracy, but the learning time and the number of candidates would likely suffer even further. In a domain with larger representations, the differences in learning time would be even more dramatic.

## 5.2 Artificial Data

The previous section showed that WOLFIE successfully learns lexicons for a natural corpus and a realistic task. However, this demonstrates success on only a relatively small corpus and with one representation formalism. We now show that our algorithm scales up well with more lexicon items to learn, more ambiguity, and more synonymy. These factors are difficult to control when using real data as input. Also, there are no large corpora available that are annotated with semantic parses. We therefore present experimental results on an artificial corpus. In this corpus, both the sentences and their representations are completely artificial, and the sentence representation is a variable-free representation, as suggested by the work of Jackendoff (1990) and others.

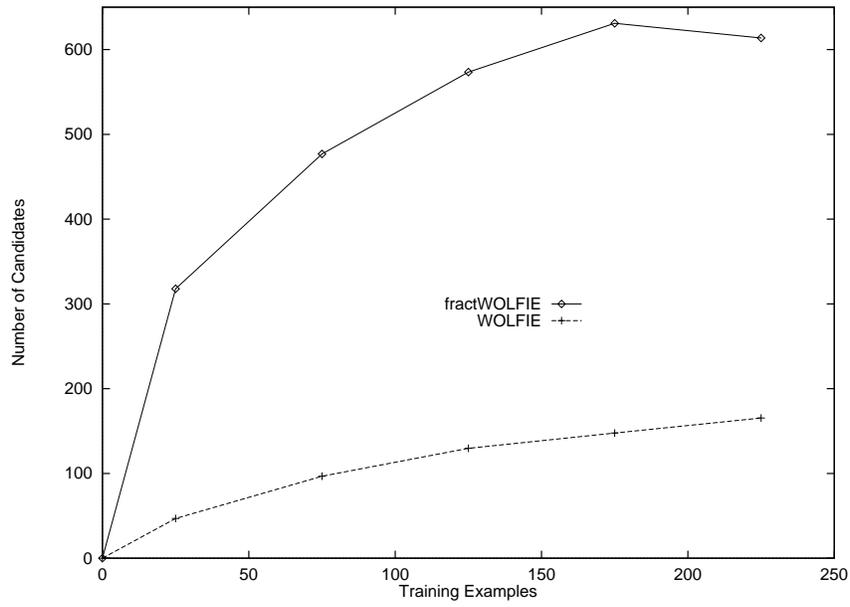

Figure 14: Fracturing vs. LICS: Number of Candidates

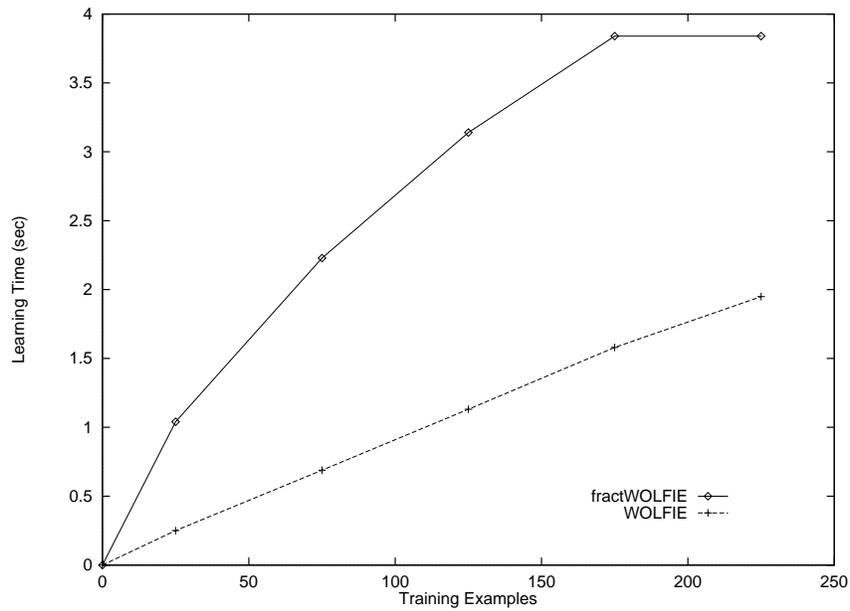

Figure 15: Fracturing vs. LICS: Learning Time

For each corpus discussed below, a random lexicon mapping words to simulated meanings was first constructed.[11] This *original* lexicon was then used to generate a corpus of random utterances each paired with a meaning representation. After using this corpus as input to WOLFIE[12], the learned lexicon was compared to the original lexicon, and *weighted precision* and *weighted recall* of the learned lexicon were measured. Precision measures the percentage of the lexicon entries (i.e., word-meaning pairs) that the system learns that are correct. Recall measures the percentage of the lexicon entries in the hand-built lexicon that are correctly learned by the system:

$$precision = \frac{\text{\# correct pairs}}{\text{\# pairs learned}}$$

$$recall = \frac{\text{\# correct pairs}}{\text{\# pairs in hand-built lexicon}}.$$

To get weighted precision and recall measures, we then weight the results for each pair by the word's frequency in the entire corpus (not just the training corpus). This models how likely we are to have learned the correct meaning for an arbitrarily chosen word in the corpus.

We generated several lexicons and associated corpora, varying the ambiguity rate (number of meanings per word) and synonymy rate (number of words per meaning), as in Siskind (1996). Meaning representations were generated using a set of "conceptual symbols" that combined to form the meaning for each word. The number of conceptual symbols used in each lexicon will be noted when we describe each corpus below. In each lexicon, 47.5% of the senses were variable-free to simulate noun-like meanings, and 47.5% contained from one to three variables to denote open argument positions to simulate verb-like meanings. The remainder of the words (the remaining 5%) had the empty meaning to simulate function words. In addition, the functors in each meaning could have a depth of up to two and an arity of up to two. An example noun-like meaning is `f23(f2(f14))`, and a verb-meaning `f10(A,f15(B))`; the conceptual symbols in this example are `f23`, `f2`, `f14`, `f10`, and `f15`. By using these multi-level meaning representations we demonstrate the learning of more complex representations than those in the geography database domain: none of the hand-built meanings for phrases in that lexicon had functors embedded in arguments. We used a grammar to generate utterances and their meanings from each original lexicon, with terminal categories selected using a distribution based on Zipf's Law (Zipf, 1949). Under Zipf's Law, the occurrence frequency of a word is inversely proportional to its ranking by occurrence.

We started with a baseline corpus generated from a lexicon of 100 words using 25 conceptual symbols and no ambiguity or synonymy; 1949 sentence-meaning pairs were generated.

---

11. Thanks to Jeff Siskind for the initial corpus generation software, which we enhanced for these tests.
12. In these tests, we allowed WOLFIE to learn phrases of up to length two.

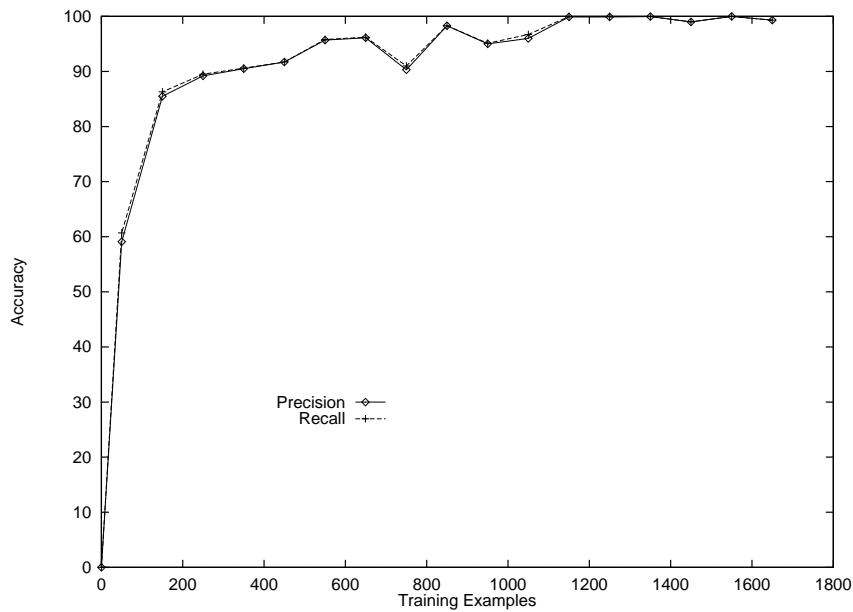

Figure 16: Baseline Artificial Corpus

We split this into five training sets of 1700 sentences each. Figure 16 shows the weighted precision and recall curves for this initial test. This demonstrates good scalability to a slightly larger corpus and lexicon than that of the U.S. geography query domain.

A second corpus was generated from a second lexicon, also of 100 words using 25 conceptual symbols, but increasing the ambiguity to 1.25 meanings per word. This time, 1937 pairs were generated and the corpus split into five sets of 1700 training examples each. Weighted precision at 1650 examples drops to 65.4% from the previous level of 99.3%, and weighted recall to 58.9% from 99.3%. The full learning curve is shown in Figure 17. A quick comparison to Siskind's performance on this corpus confirmed that his system achieved comparable performance, showing that with current methods, this is close to the best performance that we are able to obtain on this more difficult corpus. One possible explanation for the smaller performance difference between the two systems on this corpus versus the geography domain is that in this domain, the correct meaning for a word is not necessarily the most "general," in terms of number of vertices, of all its candidate meanings. Therefore, the generality portion of the heuristic may negatively influence the performance of WOLFIE in this domain.

Finally, we show the change in performance with increasing ambiguity and increasing synonymy, holding the number of words and conceptual symbols constant. Figure 18 shows the weighted precision and recall with 1050 training examples for increasing levels of ambiguity, holding the synonymy level constant. Figure 19 shows the results at increasing levels of synonymy, holding ambiguity constant. Increasing the level of synonymy does not effect the results as much as increasing the level of ambiguity, which is as we expected. Holding the corpus size constant but increasing the number of competing meanings for a word increases the number of candidate meanings created by WOLFIE while decreasing the amount of evidence available for each meaning (e.g., the first component of the heuristic measure) and makes the learning task more difficult. On the other hand, increasing the level of synonymy does not have the potential to mislead the learner.

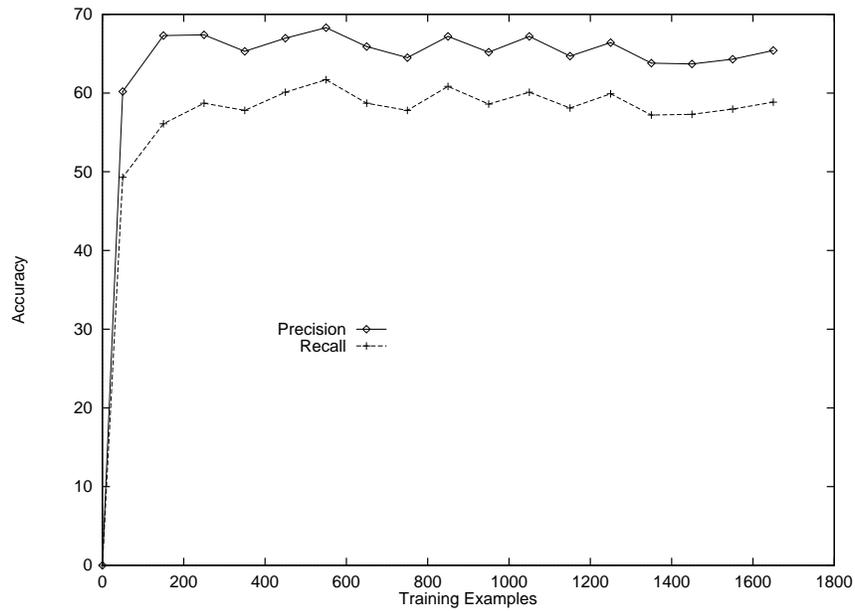

Figure 17: A More Ambiguous Artificial Corpus

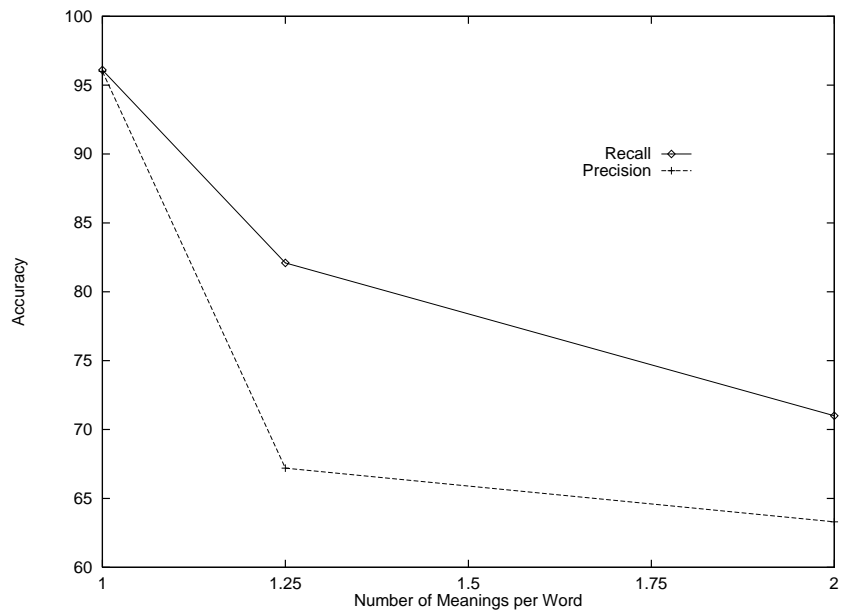

Figure 18: Increasing the Level of Ambiguity

The number of training examples required to reach a certain level of accuracy is also informative. In Table 4, we show the point at which a standard precision of 75% was first reached for each level of ambiguity. Note, however, that we only measured accuracy after each set of 100 training examples, so the numbers in the table are approximate.

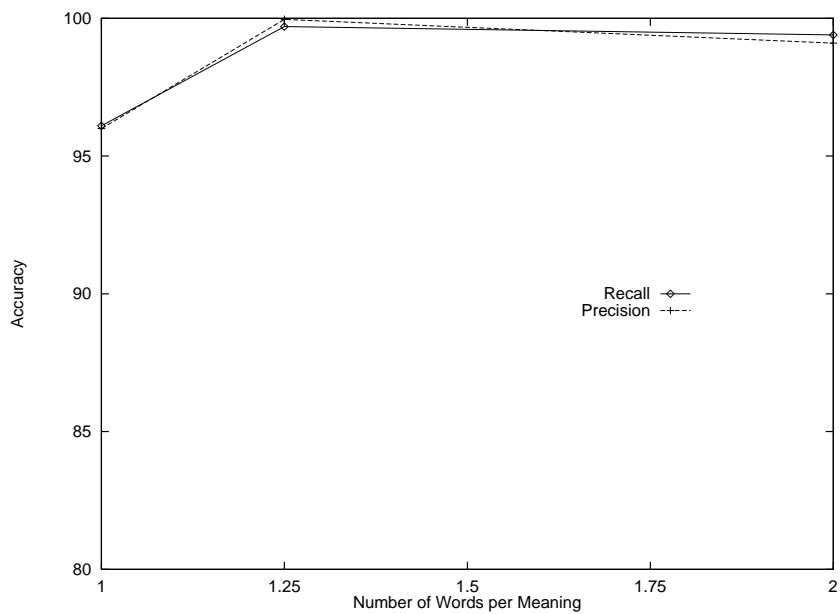

Figure 19: Increasing the Level of Synonymy

| Ambiguity Level | Number of Examples |
|---|---|
| 1.0 | 150 |
| 1.25 | 450 |
| 2.0 | 1450 |

Table 4: Number of Examples to Reach 75% Precision

We performed a second test of scalability on two corpora generated from lexicons an order of magnitude larger than those in the above tests. In these tests, we use a lexicon containing 1000 words and using 250 conceptual symbols. We generated both a corpus with no ambiguity, and one from a lexicon with ambiguity and synonymy similar to that found in the WordNet database (Beckwith, Fellbaum, Gross, & Miller, 1991); the ambiguity there is approximately 1.68 meanings per word and the synonymy 1.3 words per meaning. These corpora contained 9904 (no ambiguity) and 9948 examples, respectively, and we split the data into five sets of 9000 training examples each. For the easier large corpus, the maximum average of weighted precision and recall was 85.6%, at 8100 training examples, while for the harder corpus, the maximum average was 63.1% at 8600 training examples.

## 6. Active Learning

As indicated in the previous sections, we have built an integrated system for language acquisition that is flexible and useful. However, a major difficulty remains: the construction of training corpora. Though annotating sentences is still arguably less work than building

---

Apply the learner to $n$ bootstrap examples, creating a classifier.

**Until** no examples remain or the annotator is unwilling to label more examples, **do**:

 Use most recently learned classifier to annotate each unlabeled instance.

 Find the $k$ instances with the lowest annotation certainty.

 Annotate these instances.

 Train the learner on the bootstrap examples and all examples annotated so far.

---

Figure 20: Selective Sampling Algorithm

an entire system by hand, the annotation task is also time-consuming and error-prone. Further, the training pairs often contain redundant information. We would like to minimize the amount of annotation required while still maintaining good generalization accuracy.

To do this, we turned to methods in *active learning*. Active learning is a research area in machine learning that features systems that automatically select the most informative examples for annotation and training (Angluin, 1988; Seung, Opper, & Sompolinsky, 1992), rather than relying on a benevolent teacher or random sampling. The primary goal of active learning is to reduce the number of examples that the system is trained on, while maintaining the accuracy of the acquired information. Active learning systems may construct their own examples, request certain types of examples, or determine which of a set of unsupervised examples are most usefully labeled. The last approach, *selective sampling* (Cohn et al., 1994), is particularly attractive in natural language learning, since there is an abundance of text, and we would like to annotate only the most informative sentences. For many language learning tasks, annotation is particularly time-consuming since it requires specifying a complex output rather than just a category label, so reducing the number of training examples required can greatly increase the utility of learning.

In this section, we explore the use of active learning, specifically selective sampling, for lexicon acquisition, and demonstrate that with active learning, fewer examples are required to achieve the same accuracy obtained by training on randomly chosen examples.

The basic algorithm for selective sampling is relatively simple. Learning begins with a small pool of annotated examples and a large pool of unannotated examples, and the learner attempts to choose the most informative additional examples for annotation. Existing work in the area has emphasized two approaches, *certainty-based* methods (Lewis & Catlett, 1994), and *committee-based* methods (McCallum & Nigam, 1998; Freund, Seung, Shamir, & Tishby, 1997; Liere & Tadepalli, 1997; Dagan & Engelson, 1995; Cohn et al., 1994); we focus here on the former.

In the certainty-based paradigm, a system is trained on a small number of annotated examples to learn an initial classifier. Next, the system examines unannotated examples, and attaches certainties to the predicted annotation of those examples. The $k$ examples with the lowest certainties are then presented to the user for annotation and retraining. Many methods for attaching certainties have been used, but they typically attempt to estimate the probability that a classifier consistent with the prior training data will classify a new example correctly.

---

Learn a lexicon with the examples annotated so far
1) **For** each phrase in an unannotated sentence:
    **If** it has entries in the learned lexicon **then**
      its certainty is the average of the heuristic values of those entries
    **Else, if** it is a one-word phrase **then**
      its certainty is zero
2) To rank sentences use:
    $$\frac{\text{Total certainty of phrases from step 1}}{\text{\# of phrases counted in step 1}}$$

---

Figure 21: Active Learning for WOLFIE

Figure 20 presents abstract pseudocode for certainty-based selective sampling. In an ideal situation, the batch size, $k$, would be set to one to make the most intelligent decisions in future choices, but for efficiency reasons in retraining batch learning algorithms, it is frequently set higher. Results on a number of classification tasks have demonstrated that this general approach is effective in reducing the need for labeled examples (see citations above).

Applying certainty-based sample selection to WOLFIE requires determining the certainty of a complete annotation of a potential new training example, despite the fact that individual learned lexical entries and parsing operators perform only part of the overall annotation task. Therefore, our general approach is to compute certainties for pieces of an example, in our case, phrases, and combine these to obtain an overall certainty for an example. Since lexicon entries contain no explicit uncertainty parameters, we used WOLFIE's heuristic measure to estimate uncertainty.

To choose the sentences to be annotated in each round, we first bootstrapped an initial lexicon from a small corpus, keeping track of the heuristic values of the learned items. Then, for each unannotated sentence, we took an average of the heuristic values of the lexicon entries learned for phrases in that sentence, giving a value of zero to unknown words but eliminating from consideration any words that we assume are known in advance, such as database constants. Thus, longer sentences with only a few known phrases would have a lower certainty than shorter sentences with the same number of known phrases; this is desirable since longer sentences will be more informative from a lexicon learning point of view. The sentences with the lowest values were chosen for annotation, added to the bootstrap corpus, and a new lexicon learned. Our technique is summarized in Figure 21.

To evaluate our technique, we compared active learning to learning from randomly selected examples, again measuring the effectiveness of learned lexicons as background knowledge for CHILL. We again used the (smaller) U.S. Geography corpus, as in the original WOLFIE tests, using the lexicons as background knowledge during parser acquisition (and using the *same* examples for parser acquisition).

For each trial in the following experiments, we first randomly divide the data into a training and test set. Then, $n = 25$ bootstrap examples are randomly selected from the training examples and in each step of active learning, the least certain $k = 10$ examples of the remaining training examples are selected and added to the training set. The result of learning on this set is evaluated after each step. The accuracy of the resulting learned parsers was compared to the accuracy of those learned using randomly chosen examples to learn lexicons and parsers, as in Section 5; in other words, we can think of the $k$ examples in each round as being chosen randomly.

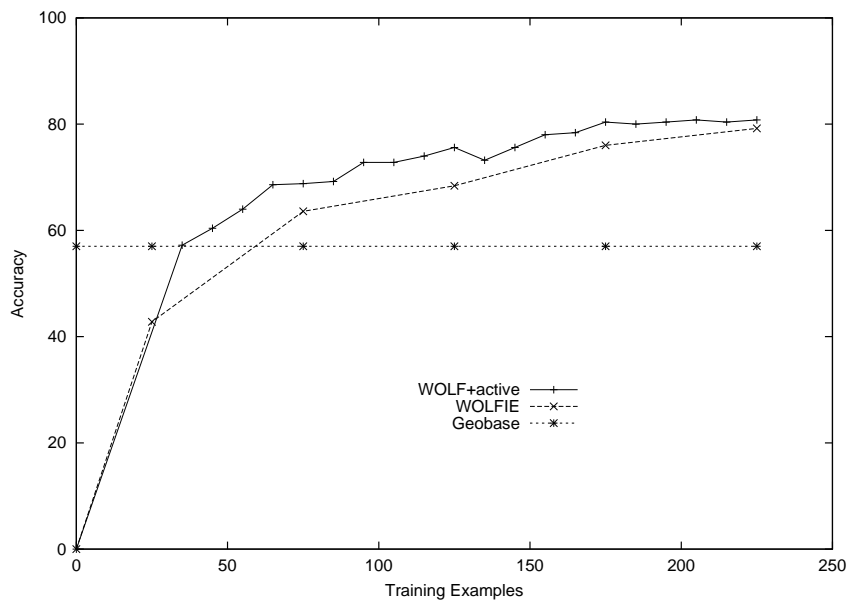

Figure 22: Using Lexicon Certainty for Active Learning

Figure 22 shows the accuracy on unseen data of parsers learned using the lexicons learned by WOLFIE when examples are chosen randomly and actively. There is an annotation savings of around 50 examples by using active learning: the maximum accuracy is reached after 175 examples, versus 225 with random examples. The advantage of using active learning is clear from the beginning, though the differences between the two curves are only statistically significant at 175 training examples. Since we are learning both lexicons and parsers, but only choosing examples based on WOLFIE's certainty measures, the boost could be improved even further if CHILL had a say in the examples chosen. See Thompson, Califf, and Mooney (1999) for a description of active learning for CHILL.

## 7. Related Work

In this section, we divide the previous research on related topics into the areas of lexicon acquisition and active learning.

### 7.1 Lexicon Acquisition

Work on automated lexicon and language acquisition dates back to Siklossy (1972), who demonstrated a system that learned transformation patterns from logic back to natural language. As already noted, the most closely related work is that of Jeff Siskind, which we described briefly in Section 2 and whose system we ran comparisons to in Section 5. Our definition of the learning problem can be compared to his "mapping problem" (Siskind, 1993). That formulation differs from ours in several respects. First, his sentence representations are terms instead of trees. However, as shown in Figure 7, terms can also be represented as trees that conform to our formalism with some minor additions. Next, his notion of interpretation does involve a type of tree, but carries the entire representation of a sentence up to the root. Also, it is not clear how he would handle quantified variables in the representation of sentences. Skolemization is possible, but then generalization across sentences would require special handling. We make the single-use assumption and he does not. Another difference is our bias towards a minimal number of lexicon entries, while he attempts to find a monosemous lexicon. His later work (Siskind, 2000) relaxes this to allow ambiguity and noise, but still biases towards minimizing ambiguity. However, his formal definition does not explicitly allow lexical ambiguity, but handles it in a heuristic manner. This, though, may lead to more robustness than our method in the face of noise. Finally, our definition allows phrasal lexicon entries.

Siskind's work on this topic has explored many different variations along a continuum of using many constraints but requiring more time to incorporate each new example (Siskind, 1993), versus few constraints but requiring more training data (Siskind, 1996). Thus, perhaps his earlier systems would have been able to learn the lexicons of Section 5 more quickly; but crucially those systems did not allow lexical ambiguity, and thus also may not have learned as accurate a lexicon. More detailed comparisons to such versions of the system are outside the scope of this paper. Our goal with WOLFIE is to learn a possibly ambiguous lexicon from as few examples as possible, and we thus made comparisons along this dimension alone.

Siskind's approach, like ours, takes into account constraints between word meanings that are justified by the exclusivity and compositionality assumptions. His approach is somewhat more general in that it handles noise and referential uncertainty (uncertainty about the meaning of a sentence and thus multiple possible candidates), while ours is specialized for applications where the meaning (or meanings) is known. The experimental results in Section 5 demonstrate the advantage of our method for such an application. He has demonstrated his system to be capable of learning reasonably accurate lexicons from large, ambiguous, and noisy artificial corpora, but this accuracy is only assured if the learning algorithm converges, which did not occur for our smaller corpus in the experiments we ran. Also, as already noted, his system operates in an incremental or on-line fashion, discarding each sentence as it processes it, while ours is batch. In addition, his search for word meanings proceeds in two stages, as discussed in Section 2.2. By using common substructures, we combine these two stages in WOLFIE. Both systems do have greedy aspects, ours in the choice of the next best lexical entry, his in the choice to discard utterances as noise or create a homonymous lexical entry. Finally, his system does not compute statistical correlations between words and their possible meanings, while ours does.

Besides Siskind's work, there are others who approach the problem from a cognitive perspective. For example, De Marcken (1994) also uses child language learning as a motivation, but approaches the segmentation problem instead of the learning of semantics. For training input, he uses a flat list of tokens for semantic representations, but does not segment sentences into words. He uses a variant of expectation-maximization (Dempster, Laird, & Rubin, 1977), together with a form of parsing and dictionary matching techniques, to segment the sentences and associate the segments with their most likely meaning. On the Childes corpus, the algorithm achieves very high precision, but recall is not provided.

Others taking the cognitive approach demonstrate language understanding by the ability to carry out some task such as parsing. For example, Nenov and Dyer (1994) describe a neural network model to map between visual and verbal-motor commands, and Colunga and Gasser (1998) use neural network modeling techniques for learning spatial concepts. Feldman and his colleagues at Berkeley (Feldman, Lakoff, & Shastri, 1995) are actively pursuing cognitive models of the acquisition of semantic concepts. Another Berkeley effort, the system by Regier (1996) is given examples of pictures paired with natural language descriptions that apply to the picture, and learns to judge whether a new sentence is true of a given picture.

Similar work by Suppes, Liang, and Böttner (1991) uses robots to demonstrate lexicon learning. A robot is trained on cognitive and perceptual concepts and their associated actions, and learns to execute simple commands. Along similar lines, Tishby and Gorin (1994) have a system that learns associations between words and actions, but they use a statistical framework to learn these associations, and do not handle structured representations. Similarly, Oates, Eyler-Walker, and Cohen (1999) discuss the acquisition of lexical hierarchies and their associated meaning as defined by the sensory environment of a robot.

The problem of automatic construction of translation lexicons (Smadja, McKeown, & Hatzivassiloglou, 1996; Melamed, 1995; Wu & Xia, 1995; Kumano & Hirakawa, 1994; Catizone, Russell, & Warwick, 1993; Gale & Church, 1991; Brown & et al., 1990) has a definition similar to our own. While most of these methods also compute association scores between pairs (in their case, word-word pairs) and use a greedy algorithm to choose the best translation(s) for each word, they do not take advantage of the constraints between pairs. One exception is Melamed (2000); however, his approach does not allow for phrases in the lexicon or for synonymy within one text segment, while ours does. Also, Yamazaki, Pazzani, and Merz (1995) learn both translation rules and semantic hierarchies from parsed parallel sentences in Japanese and English. Of course, the main difference between this body of work and this paper is that we map words to *semantic structures*, not to other words.

As mentioned in the introduction, there is also a large body of work on learning lexical semantics but using different problem formulations than our own. For example, Collins and Singer (1999), Riloff and Jones (1999), Roark and Charniak (1998), and Schneider (1998) define semantic lexicons as a grouping of words into semantic categories, and in the latter case, add relational information. The result is typically applied as a semantic lexicon for information extraction or entity tagging. Pedersen and Chen (1995) describe a method for acquiring syntactic and semantic features of an unknown word, assuming access to an initial concept hierarchy, but they give no experimental results. Many systems (Fukumoto & Tsujii, 1995; Haruno, 1995; Johnston, Boguraev, & Pustejovsky, 1995; Webster & Marcus, 1995) focus only on acquisition of verbs or nouns, rather than all types of words. Also, the authors just named either do not experimentally evaluate their systems, or do not show the usefulness of the learned lexicons for a specific application.

Several authors (Rooth, Riezler, Prescher, Carroll, & Beil, 1999; Collins, 1997; Ribas, 1994; Manning, 1993; Resnik, 1993; Brent, 1991) discuss the acquisition of subcategoriza-tion information for verbs, and others describe work on learning selectional restrictions (Manning, 1993; Brent, 1991). Both of these are different from the information required for mapping to semantic representation, but could be useful as a source of information to further constrain the search. Li (1998) further expands on the subcategorization work by inducing clustering information. Finally, several systems (Knight, 1996; Hastings, 1996; Russell, 1993) learn new words from context, assuming that a large initial lexicon and parsing system are already available.

Another related body of work is grammar acquisition, especially those areas that tightly integrate the grammar with a lexicon, such as with Categorial Grammars (Retore & Bonato, 2001; Dudau-Sofronie, Tellier, & Tommasi, 2001; Watkinson & Manandhar, 1999). The theory of Categorial Grammar also has ties with lexical semantics, but these semantics have not often been used for inference in support of high-level tasks such as database retrieval. While learning syntax and semantics together is arguably a more difficult task, the aforementioned work has not been evaluated on large corpora, presumably primarily due to the difficulty of annotation.

## 7.2 Active Learning

With respect to additional active learning techniques, Cohn et al. (1994) were among the first to discuss certainty-based active learning methods in detail. They focus on a neural network approach to active learning in a version-space of concepts. Only a few of the researchers applying machine learning to natural language processing have utilized active learning (Hwa, 2001; Schohn & Cohn, 2000; Tong & Koller, 2000; Thompson et al., 1999; Argamon-Engelson & Dagan, 1999; Liere & Tadepalli, 1997; Lewis & Catlett, 1994), and the majority of these have addressed classification tasks such as part of speech tagging and text categorization. For example, Liere and Tadepalli (1997) apply active learning with committees to the problem of text categorization. They show improvements with active learning similar to those that we obtain, but use a committee of Winnow-based learners on a traditional classification task. Argamon-Engelson and Dagan (1999) also apply committee-based learning to part-of-speech tagging. In their work, a committee of hidden Markov models is used to select examples for annotation. Lewis and Catlett (1994) use *heterogeneous* certainty-based methods, in which a simple classifier is used to select examples that are then annotated and presented to a more powerful classifier.

However, many language learning tasks require annotating natural language text with a complex output, such as a parse tree, semantic representation, or filled template. The application of active learning to tasks requiring such complex outputs has not been well studied, the exceptions being Hwa (2001), Soderland (1999), Thompson et al. (1999). The latter two include work on active learning applied to information extraction, and Thompson et al. (1999) includes work on active learning for semantic parsing. Hwa (2001) describes an interesting method for evaluating a statistical parser's uncertainty, when applied for syntactic parsing.

## 8. Future Work

Although WOLFIE's current greedy search method has performed quite well, a better search heuristic or alternative search strategy could result in improvements. We should also more thoroughly evaluate Wolfie's ability to learn long phrases, as we restricted this ability in the evaluations here. Another issue is robustness in the face of noise. The current algorithm is not guaranteed to learn a correct lexicon in even a noise-free corpus. The addition of noise complicates an analysis of circumstances in which mistakes are likely to happen. Further theoretical and empirical analysis of these issues is warranted.

Referential uncertainty could be handled, with an increase in complexity, by forming LICS from more pairs of representations with which a phrase appears, but not between alternative representations of the same sentence. Then, once a pair is added to the lexicon, for each sentence containing that word, representations can be eliminated if they do not contain the learned meaning, provided another representation does contain it (thus allowing for lexical ambiguity). We plan to flesh this out and evaluate the results.

A different avenue of exploration is to apply Wolfie to a corpus of sentences paired with the more common query language, SQL. Such corpora should be easily constructible by recording queries submitted to existing SQL applications along with their English forms, or translating existing lists of SQL queries into English (presumably an easier direction to translate). The fact that the same training data can be used to learn both a semantic lexicon and a parser also helps limit the overall burden of constructing a complete natural language interface.

With respect to active learning, experiments on additional corpora are needed to test the ability of our approach to reduce annotation costs in a variety of domains. It would also be interesting to explore active learning for other natural language processing problems such as syntactic parsing, word-sense disambiguation, and machine translation.

Our current results have involved a certainty-based approach; however, proponents of committee-based approaches have convincing arguments for their theoretical advantages. Our initial attempts at adapting committee-based approaches to our systems were not very successful; however, additional research on this topic is indicated. One critical problem is obtaining diverse committees that properly sample the version space (Cohn et al., 1994).

## 9. Conclusions

Acquiring a semantic lexicon from a corpus of sentences labeled with representations of their meaning is an important problem that has not been widely studied. We present both a formalism of the learning problem and a greedy algorithm to find an approximate solution to it. Wolfie demonstrates that a fairly simple, greedy, symbolic learning algorithm performs well on this task and obtains performance superior to a previous lexicon acquisition system on a corpus of geography queries. Our results also demonstrate that our methods extend to a variety of natural languages besides English, and that they scale fairly well to larger, more difficult corpora.

Active learning is a new area of machine learning that has been almost exclusively applied to classification tasks. We have demonstrated its successful application to more complex natural language mappings from phrases to semantic meanings, supporting the acquisition of lexicons and parsers. The wealth of unannotated natural language data, along with the difficulty of annotating such data, make selective sampling a potentially invaluable technique for natural language learning. Our results on realistic corpora indicate that example annotations savings as high as 22% can be achieved by employing active sample selection using only simple certainty measures for predictions on unannotated data. Improved sample selection methods and applications to other important language problems hold the promise of continued progress in using machine learning to construct effective natural language processing systems.

Most experiments in corpus-based natural language have presented results on some subtask of natural language, and there are few results on whether the learned subsystems can be successfully integrated to build a complete NLP system. The experiments presented in this paper demonstrated how two learning systems, Wolfie and Chill, were successfully integrated to learn a complete NLP system for parsing database queries into executable logical form given only a single corpus of annotated queries, and further demonstrated the potential of active learning to reduce the annotation effort for learning for NLP.

## Acknowledgments

We would like to thank Jeff Siskind for providing us with his software, and for all his help in adapting it for use with our corpus. Thanks also to Agapito Sustaita, Esra Erdem, and Marshall Mayberry for their translation efforts, and to the three anonymous reviewers for their comments which helped improve the paper. This research was supported by the National Science Foundation under grants IRI-9310819 and IRI-9704943.